\documentclass{article}

    \PassOptionsToPackage{numbers, compress}{natbib}

\usepackage[preprint]{neurips_2024}

\usepackage[utf8]{inputenc} 
\usepackage[T1]{fontenc}    
\usepackage{hyperref}       
\usepackage{url}            
\usepackage{booktabs}       
\usepackage{amsfonts}       
\usepackage{nicefrac}       
\usepackage{microtype}      
\usepackage{xcolor}         

\usepackage{multirow}
\usepackage{colortbl}
\usepackage{wrapfig}
\makeatletter
\@namedef{ver@everyshi.sty}{}
\makeatother

\usepackage{colortbl}
\usepackage{graphicx}
\usepackage{subcaption}
\usepackage{lipsum}
\usepackage{arydshln}

\newcommand{\best}{\cellcolor{tablered}}
\newcommand{\sbest}{\cellcolor{orange}}
\newcommand{\tbest}{\cellcolor{yellow}}

\newcommand{\bc}{\mathbf{c}}

\newcommand{\bn}{\mathbf{n}}

\newcommand{\bt}{\mathbf{t}}

\makeatletter
\DeclareRobustCommand\onedot{\futurelet\@let@token\@onedot}
\def\@onedot{\ifx\@let@token.\else.\null\fi\xspace}

\makeatother

\definecolor{yellow}{rgb}{1, 1, 0.7}
\definecolor{orange}{rgb}{1, 0.85, 0.7}
\definecolor{tablered}{rgb}{1, 0.7, 0.7}
\definecolor{red}{rgb}{1, 0, 0}

\definecolor{wincolor}{rgb}{0.85, 0.0, 0.0}

\definecolor{darkyellow}{rgb}{0.8, 0.8, 0.5}
\definecolor{darkred}{rgb}{0.7, 0.3, 0.3}
\definecolor{darkgreen}{rgb}{0.3, 0.7, 0.3}
\definecolor{green}{rgb}{0, 1.0, 0}
\definecolor{pink}{rgb}{1, 0.4, 0.7}

\definecolor{realred}{rgb}{0.95, 0.1, 0.0}

\usepackage{amsmath}
\usepackage{extarrows}
\usepackage{enumitem}
\usepackage[misc]{ifsym}

\definecolor{cite_blue}{RGB}{38,109,185}
\definecolor{cvprblue}{rgb}{0.21,0.49,0.74}
\hypersetup{
    colorlinks=true,
    citecolor=cvprblue
}

\title{Trim 3D Gaussian Splatting for Accurate Geometry Representation}

\author{%
  Lue Fan$^{1,2}$\thanks{Equal contribution.}\quad Yuxue Yang$^{1*}$\quad Minxing Li$^1$\quad Hongsheng Li$^{2,3}\textsuperscript{\Letter}$\quad Zhaoxiang Zhang$^{1}\textsuperscript{\Letter}$\\
  \vspace{-1mm}\\
  $^1$CASIA\quad $^2$MMLab, CUHK\quad 
  $^3$Shanghai AI Lab\\
  \vspace{-1mm}\\
  \url{https://trimgs.github.io}
}

\begin{document}

\maketitle

\begin{abstract}
In this paper, we introduce Trim 3D Gaussian Splatting (TrimGS) to reconstruct accurate 3D geometry from images.
Previous arts for geometry reconstruction from 3D Gaussians mainly focus on exploring strong geometry regularization. 
Instead, from a fresh perspective, we propose to obtain accurate 3D geometry of a scene by Gaussian trimming, which selectively removes the inaccurate geometry while preserving accurate structures.
To achieve this, we analyze the contributions of individual 3D Gaussians and propose a contribution-based trimming strategy to remove the redundant or inaccurate Gaussians.
Furthermore, our experimental and theoretical analyses reveal that a relatively small Gaussian scale is a non-negligible factor in representing and optimizing the intricate details.
Therefore the proposed TrimGS maintains relatively small Gaussian scales.
In addition, TrimGS is also compatible with the effective geometry regularization strategies in previous arts.
When combined with the original 3DGS and the state-of-the-art 2DGS, TrimGS consistently yields more accurate geometry and higher perceptual quality.
\end{abstract}

\begin{figure}[h]
    \centering
    \vspace{-2mm}
    \includegraphics[width=\linewidth]{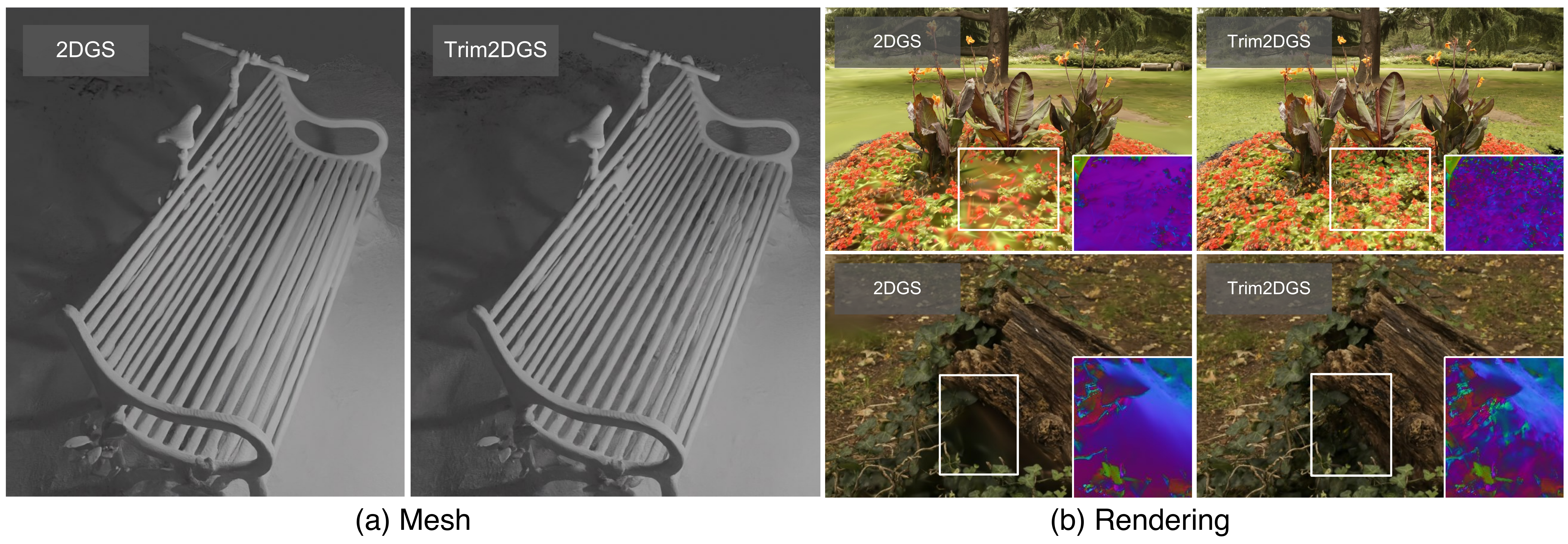}
    \vspace{-6mm}
    \caption{\textbf{TrimGS exhibits better geometric details and perceptual quality.} In (a), TrimGS separates the slender crossbars of a bench. (b) showcases that TrimGS excels in rendering intricate details in both color and normals.}
    \vspace{-5mm}
    \label{fig:properties}
\end{figure}
\section{Introduction}
\label{sec:intro}

In novel view synthesis, it is a common phenomenon that messy geometry might be hidden behind the high rendering quality, as highlighted by NeRF++~\cite{zhang2020nerf++}.
This issue becomes more prominent with the explicit representation of 3D Gaussian Splatting (3DGS)~\cite{3dgs}.
To derive accurate explicit geometry from 3D Gaussian representation, recent methods~\cite{sugar, 2dgs, gof, GaussianSurfels, gs2mesh} have made attempts to reconstruct accurate surfaces from 3D Gaussians.
Despite differing methodological focuses, these approaches share the strategy of applying strong geometric regularization to the original 3DGS.

In this paper, we propose to \emph{\textbf{trim}} 3D Gaussians, which is a new strategy for accurate geometry reconstruction that complements the previous geometry regularization strategies.
Unlike the term ``pruning'' used in 3DGS, our proposed gradual ``trimming'' strategy is employed to highlight our method's goal of progressively refining messy Gaussian fields into geometrically accurate forms, akin to lawn trimming.
To this end, we first introduce a metric to determine which parts of the Gaussians should be trimmed. In the original 3DGS, opacity serves as this metric, with Gaussians exhibiting very low opacities being removed.
However, many Gaussians with low contributions or inaccurate geometric structures often have relatively high opacities, making the default strategy insufficient for maintaining precise geometry. To address this issue, we develop a faithful evaluation metric for measuring Gaussians' contributions, inspired by the weights used in the alpha-blending process. We then design an intuitive yet effective contribution-based trimming procedure and integrate it into the 3DGS training process.

In evaluating the contributions, the scale of Gaussians is a non-negligible factor.
Although large Gaussians often have higher contributions to the rendering process, their large sizes limit their ability to represent intricate geometry.
They tend to produce blurred patterns in high-frequency regions, significantly diminishing perceptual quality.
Beyond their limited capacity, our experimental and theoretical analyses, detailed in \S~\ref{sec:pre} and \S~\ref{sec:appendix_gradient}, reveal that the optimization of large Gaussians is hampered by noisy gradients. To address these drawbacks, we propose modifying the densification strategy in 3DGS to control and maintain relatively small Gaussian scales.

Although we follow a new perspective, the proposed TrimGS remains compatible with existing geometry regularization strategies.
Therefore, we also incorporate a normal consistency regularization to ensure better and more stable geometry optimization.
Furthermore, the compatibility makes TrimGS seamlessly integrated with the emerging 2DGS.

Our contributions are summarized as follows.

\begin{itemize}
    \item We propose TrimGS to learn 3D Gaussians with accurate geometry, utilizing a contribution-based Gaussian trimming strategy.
    It offers a fresh perspective on geometry reconstruction from 3D Gaussians, complementing conventional geometry regularization techniques.
    \item Through experimental and theoretical analysis, we highlight Gaussian scale is a non-negligible factor in trimming and optimization. Thus, we advocate for maintaining relatively small Gaussian scales, with the hope that our analysis provides insights to the community.
    \item The proposed TrimGS is a general technique and compatible with the original 3DGS and the recent 2DGS methods.
    Based on them, TrimGS consistently produces more accurate geometry and higher perceptual quality.
\end{itemize}

\section{Related work}

\subsection{Novel View Synthesis}

Methods for Novel View Synthesis (NVS) can be partitioned into implicit methods and explicit methods based on the used representation.
NeRF~\cite{mildenhall2020nerf} was the first to employ implicit radiance fields for representing 3D scenes for NVS. This pioneering work uses a multi-layer perceptron (MLP) to encode geometric and photometric information at every spatial position. The directional encoding enables view-dependent appearances. Volume rendering techniques~\cite{kajiya1984ray, drebin1988volume} are leveraged to render the implicit neural field into images.
The most significant issue for the original NeRF is the low training and rendering efficiency.
Thus a line of methods aims for improving its efficiency.
Grid-based feature representation is proven effective in accelerating convergence.
Representative methods~\cite{muller2022instant, kulhanek2023tetranerf, yu2022plenoxels, Sun2022CVPR} propose to optimize the grid-based feature instead of neural parameters in MLP.
For the rendering efficiency, baking strategy~\cite{Reiser2021ICCV, Hedman2021ICCV, yariv2023bakedsdf, reiser2023merf} and sparse hash-based grid~\cite{muller2022instant} are leveraged for performance boosting.
Another line of works~\cite{barron2022mip, Barron2023ICCV, hu2023tri} focuses on improving the rendering quality of NeRF, especially addressing the anti-aliasing issue.
On the other hand, explicit radiance fields, such as Plenoxels~\cite{yu2022plenoxels}, and more recently, 3D Gaussian Splatting~\cite{3dgs}, offer alternative approaches and are getting popular.
The 3DGS employs 3D Gaussians to depict complex scenes, achieving impressive rendering quality and efficiency.
However, the reconstructed geometry of the original 3DGS is quite messy. In the following paragraph, we briefly introduce the recent advances in accurate geometry extraction from 3DGS.

\subsection{Surface Reconstruction with Gaussians}
The explicit representation of 3D Gaussians makes extracting accurate geometry an intriguing topic. 
Almost all previous methods~\cite{sugar, Turkulainen2024ARXIV, 2dgs, chen2023neusg, Yu2024GSDF},  focus on applying geometric regularization, which is proven effective.
Notably, SuGaR~\cite{sugar} leverages signed distance function and density to supervise Gaussian distribution, forcing them to align with object surfaces.
Then it extracts meshes with Poisson surface reconstruction~\cite{kazhdan2013screened}.
Finally, it proposes a novel binding strategy to bind 3D Gaussians with coarse meshes and conduct further refinement.
SuGaR enables flexible editing of 3D Gaussians by editing the coupled meshes.
Recently, NeuSG~\cite{chen2023neusg} and GSDF~\cite{Yu2024GSDF} combine 3D Gaussians with implicit surfaces.
They optimize 3D Gaussian fields together with signed distance functions to obtain high-quality surfaces.
GS2Mesh~\cite{gs2mesh} utilizes stereo depth estimation to enhance the depth quality for fusion.
The most recent 2DGS~\cite{2dgs} introduces 2D disk representation to replace the 3D representation, yielding more smooth object surfaces.
Meanwhile, in addition to geometry regularization, 2DGS employs Truncated Signed Distance Function (TSDF) fusion and Marching Cubes~\cite{lorensen1998marching} for mesh extraction, which shows impressive robustness to floaters and noisy depth.
A similar idea of combining Gaussians and surfels is concurrently proposed by GaussianSurfels~\cite{GaussianSurfels}.
GOF~\cite{gof}, a concurrent work of 2DGS, mainly focuses on unbounded scenes. It leverages ray-tracing-based volume rendering of 3D Gaussians, enabling directly extracting geometry 3D Gaussians utilizing levelset, without resorting to Poisson reconstruction or TSDF fusion as in SuGaR and 2DGS.
Although these methods produce smooth surfaces with strong geometric regularization, they face challenges in capturing detailed geometry and textures. 
In this paper, in addition to solely focusing on regularization, our method learns geometrically accurate 3D Gaussian by trimming strategy and enhancing the detailed quality by scale control.

\subsection{Pruning Strategies for Gaussians}
Pruning is an important technique in 3DGS. 
Recently, some methods prune Gaussians during training to control the total number of Gaussian primitives, whether incorporating additional modules~\cite{lee2023compact, ji2024neds} or designing a certain criterion ~\cite{fan2023light, kerbl2024reducing}. \cite{lee2023compact} develops a learnable mask strategy that eliminates Gaussian based on its volume and transparency while \cite{ji2024neds} applies a Virtual Camera View Pruning method to mark and eliminate outliers. \cite{kerbl2024reducing} calculates a redundancy score by an intersection test according to which Gaussians are filtered and \cite{fan2023light} designs a significant score that measures the contribution of a Gaussian to the rendered image and prunes those with low scores.

While these methods all focus on the pruning strategy, they mainly treat it as an approach to the reduction of memory cost~\cite{kerbl2024reducing, lee2023compact, fan2023light} instead of more accurate geometry. Therefore, they tend to keep larger Gaussians while pruning the smaller ones, which reduces the capacity to represent geometry details.


\section{Preliminaries}
\label{sec:pre}
In this section, we present the preliminaries of 3D Gaussian Splatting (3DGS) and our analysis of its properties.

\subsection{3D Gaussian Splatting}

3DGS creates a set of 3D Gaussians to explicitly represent the target 3D scene. Each Gaussian is defined by a position $\mu$, covariance matrix $\Sigma$, opacity $\sigma$, and SH coefficients that determine its view-dependent color.
To render an image, 3DGS utilizes an alpha-blending process to accumulate the color of each pixel. 
Let $p$ denote the target pixel, its color $I_p$ is given by
\begin{equation}\label{eq:alpha_blending}
    I_p = \sum_{i\in \Lambda}\bc_i\alpha_i\prod_{j=1}^{i-1}(1-\alpha_j),
\end{equation}
where $\Lambda$ is the set of Gaussians that affects $p$, with their depth in ascending order. $\bc_i$ denotes the color of the i-$th$ Gaussian generated by SH coefficients, depending on the observing point.
The blending weight $\alpha_i$ is acquired by querying from the corresponding Gaussian distribution
\begin{equation}\label{eq:gaussian_3d}
    \alpha_i = \sigma_i \cdot \exp\left(-\frac{1}{2}(\boldsymbol{p}-\boldsymbol{\mu}_i)^T\boldsymbol{\Sigma}_i^{-1}(\boldsymbol{p}-\boldsymbol{\mu}_i)\right),
\end{equation}
with $\boldsymbol{p}$ denoting the coordinate of pixel image space and $\sigma_i$ being the opacity parameters of the $i$-th Gaussian. 
It is noticeable that the normalization coefficient of Gaussian distribution is ignored in Eq.~\eqref{eq:gaussian_3d}, with the standard version given as follows
\begin{equation}\label{eq:gaussian_normal}
    G(\boldsymbol{x};\boldsymbol{\mu},\boldsymbol{\Sigma}) = \frac{1}{(2\pi)^{3/2}(\det \boldsymbol{\Sigma})^{1/2}}\exp\left(-\frac{1}{2}(\boldsymbol{x}-\boldsymbol{\mu})^T \boldsymbol{\Sigma} ^{-1} (\boldsymbol{x}-\boldsymbol{\mu})\right).
\end{equation}
Thus, the Gaussian distribution adopted in 3DGS implementation is actually an \emph{\textbf{unnormalized}} formulation.
This property is considered in the following gradient analysis and the contribution metric design in \S~\ref{sec:prune}.

\subsection{Gradient Analysis}
\label{sec:prec_grad}
As we mentioned in \S~\ref{sec:intro}, large Gaussians usually have more noisy gradients.
Here we conduct a simple analysis.
During training, we accumulate the gradient of the Gaussian positions for hundreds of iterations.
Due to the ignored normalization factor in Eq.~\eqref{eq:gaussian_3d}, we manually normalize the gradient by the determinant of the covariance $\Sigma$ for fair gradient comparison between Gaussians of different sizes.
The relationship between Gaussian sizes and gradients is shown in Table~\ref{tab:gradient_analysis}.
As can be seen, large Gaussians have much smaller normalized gradients.

\begin{table}[ht]
\centering

\caption{\textbf{The relationship between Gaussian size and gradients.} The size of a 3D Gaussian is measured by the determinant of its covariance matrix $\Sigma$. 
}
\vspace{2mm}
\begin{tabular}{l||cccc}
Size & $[10^{-6}, 10^{-5})$ & $ [10^{-5}, 10^{-4})$ & $[10^{-4}, 10^{-3})$ & $[10^{-3}, \infty)$  \\
\hline
Normalized gradient norm & 9.85 & 2.81 & 0.60 & 0.08  \\
\end{tabular}
\label{tab:gradient_analysis}
\end{table}

We attribute this to the conflict of pixel-wise gradient.
For each Gaussian, its gradient is computed by summing the gradients of pixel-wise photometric loss.
Inconsistent gradients of these pixels result in a relatively small summation.
Large Gaussians, which cover a wide range of pixels, are more likely to be trapped in this dilemma, especially when the target 3D scene is complex and rich in detail.
A theoretical demonstration of this phenomenon is given in Appendix~\ref{sec:appendix_gradient}.

\section{Method}
Our methodology has three components.
The major one is contribution-based trimming presented in \S~\ref{sec:prune}.
We further introduce a scale control strategy for better contribution evaluation and rendering details in \S~\ref{sec:split}.
Finally, we propose a normal regularization to enhance the geometry in \S~\ref{sec:normal}.

\subsection{Gaussian Trimming}
\label{sec:prune}

The most important factor in Gaussian Trimming is to properly evaluate the contribution of each individual Gaussian.
The original 3DGS directly adopts Gaussian opacity as the contribution and removes the low-opacity ones.
However, as discussed in \S~\ref{sec:intro}, such an opacity-based strategy might mistakenly removes important Gaussians but preserves high-opacity floaters.
Inspired by the alpha-blending process, we propose a more faithful metric to evaluate the Gaussian contribution.

\paragraph{Single-view Contribution.}
We first consider the contribution of a Gaussian to a single view.
In the alpha blending formulated by Eq.~\eqref{eq:alpha_blending}, the blending weight $\alpha_i\prod_{j=1}^{i-1}(1-\alpha_j)$ can be used to indicate the Gaussian's contribution to a pixel.
In this way, the overall contribution $\mathbf{C}_k$ of a Gaussian to the $k$-th rendered image can be the sum of alpha-blending weights of all related pixels, formulated as
\begin{equation}
    \mathbf{C}_k = \sum_{p \in \mathbb{P}_{k}}\alpha_{i(p)}\prod_{j=1}^{i(p)-1}(1-\alpha_{j}).
    \label{eq:contrib}
\end{equation}
In Eq.~\eqref{eq:contrib}, $\mathbb{P}_{k}$ stands for the 2D projected region of the Gaussian in k-$th$ view. Here we use the notation $i(p)$ instead of $i$ used in Eq.~\eqref{eq:alpha_blending} because the Gaussian has different depth orders in different pixel rays.

Eq.~\eqref{eq:contrib} is the theoretically accurate evaluation of Gaussian contribution. 
However, according to our analysis in \S~\ref{sec:pre}, the Gaussian distribution adopted in 3DGS is not normalized.
Directly applying Eq.~\eqref{eq:contrib} outputs quite significant contributions for large Gaussians and very low contributions for small Gaussians.
Since large Gaussians have limited capacities to represent the details and are hard to optimize as demonstrated in \S~\ref{sec:prec_grad}, we further manually normalize Eq.~\eqref{eq:contrib} into
\begin{equation}
    \mathbf{C}_k = \frac{1}{|\mathbb{P}_{k}|}\sum_{p \in \mathbb{P}_{k}}\left(\alpha_{i(p)}\right)^{\gamma}\left(\prod_{j=1}^{i(p)-1}(1-\alpha_{j})\right)^{(1 - \gamma)}.
    \label{eq:contrib2}
\end{equation}
Eq.~\eqref{eq:contrib2} has two differences from Eq.~\eqref{eq:contrib}:
(1) The contribution is normalized by the number of pixels in the projected area;
(2) We introduce a hyper-parameter $\gamma$ to control the contribution components. 
When $\gamma$ becomes larger, Eq.~\eqref{eq:contrib2} has less weight for the transmittance $\prod_{j=1}^{i(p)-1}(1-\alpha_{j})$ and degrades to the default opacity-based pruning in 3DGS.

Another reason for introducing $\gamma$ is to control a contribution bias. Intuitively, points deviating from the surface usually lead to inaccurate geometry.
Among these inaccurate Gaussians,  those near outside surfaces usually have large transmittance while inside ones usually have smaller transmittance.
Therefore, $\gamma$ controls the bias toward the Gaussians deviating to inner sides or outer sides.

\paragraph{Multi-view Contribution.}
Then we consider the overall contribution of a Gaussian to all views.
A Gaussian usually has large contributions only in a limited number of views. Thus, we use the average contribution value of a small number of high-contribution views as the overall contribution, formally denoted as
\begin{equation}
    \mathbf{C} = \frac{1}{|\mathbb{V}|}\sum_{k \in \mathbb{V}}\mathbf{C}_k.
    \label{eq:overall_contrib}
\end{equation}
$\mathbb{V}$ is the view set where the Gaussian has top contributions and typically we choose top-5 views.
In practice, the contribution evaluation method can be efficiently implemented with some simple modifications in the CUDA kernel.

\paragraph{Trimming.}
During training, we perform the proposed trimming at pre-defined intervals of iterations, where the interval is typically 1,000 iterations.
At each time, we traverse all views in the training set to evaluate the multi-view contribution of each Gaussian.
A certain percentage of Gaussians with the lowest contributions are removed.

\subsection{Scale-driven Densification}
\label{sec:split}

According to our analysis in \S~\ref{sec:prec_grad}, large Gaussian is a sub-optimal representation for geometry details and high-frequency regions.
To maintain small Gaussian sizes, if the maximum scale of a Gaussian is larger than a scene-dependent scale threshold $\tau_s$, we split it into $K$ smaller Gaussians and shrink the scale by a predefined factor, similar to the splitting implementation in the original 3DGS.

This strategy essentially abides by gradient-driven densification in the original 3DGS.
If a Gaussian gradually grows to a large Gaussian, it is likely to receive significant \textbf{accumulated} gradients for scales.
Thus, splitting those large Gaussians has a similar effect to splitting the Gaussians with large gradients while being more straightforward and intuitive.

\subsection{Normal Regularization}
\label{sec:normal}
In addition to Gaussian trimming, TrimGS can also be seamlessly combined with geometric regularization in previous methods.
We therefore further propose a normal regularization loss for better geometry learning. 
The basic idea is forcing the consistency between the normals of Gaussians and the normals derived from rendered depth maps.
The normal of a Gaussian is defined as the orientation of its shortest axis.
The challenge here is that both Gaussian normals and depth-derived normals are quite noisy at the beginning, leading to noisy supervision. 
Thus we propose a robust normal calculation method from rendered depth maps.
\begin{wrapfigure}{r}{0.4\textwidth}
    \centering
    \includegraphics[width=\linewidth]{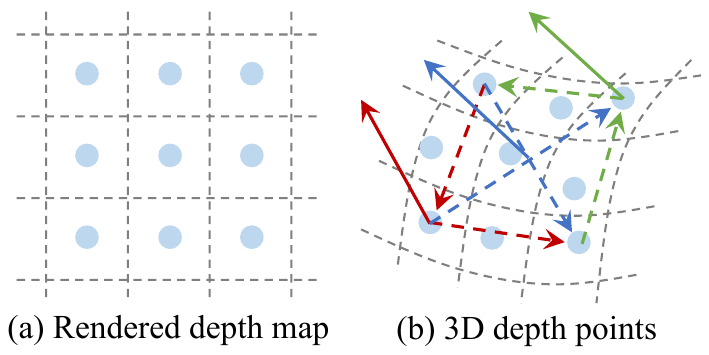}
    \caption{\textbf{Illustration of robust normal calculation from rendered depth map.}}
    \vspace{-6mm}
\label{fig:normal_reg}
\end{wrapfigure}
Considering a local $k \times k$ window in a rendered depth map, there are $k^2$ depth points $p \in \mathbb{R}^3$, which can be calculated from the depth value and the camera pose.
Any pair of the $k^2$ depth points represents a 3D vector in the tangent plane of the surface.
Thus a normal can be derived from any two pairs of depth points by cross-product.
We formally define the normal of the local window as 
\begin{equation}
    \bn = \frac{1}{|\mathbb{T}|}\sum_{\bt_i, \bt_j \in \mathbb{T}} \bt_i \times \bt_j,
\end{equation}
where $\mathbb{T}$ is the set of all tangent vectors defined by pairs of depth points in the local window. 
In practice, we do not traverse all the tangent vectors in $\mathbb{T}$ but sample some of them as Figure~\ref{fig:normal_reg} (b) shows.
This normal map derived from a depth map is denoted as $\mathbf{N}_D$.

On the other hand, we render Gaussian normals into another normal map $\mathbf{N}_G$. L1 loss between $\mathbf{N}_D$ and $\mathbf{N}_G$ is adopted to ensure the normal consistency. In this way, the overall loss function is
\begin{equation}
    \mathcal{L} = \alpha_1 \mathcal{L}_c + \alpha_2 |\mathbf{N}_D - \mathbf{N}_G|_1,
    \label{eq:loss}
\end{equation}
where the $\mathcal{L}_c$ is the default photometric loss in 3DGS. For a fair comparison, when combined with 2DGS, it utilizes the depth distortion loss and the default normal regularization in 2DGS.

\section{Experiments}

\subsection{Setup}
\paragraph{Model variants.} We apply TrimGS to both original 3DGS~\cite{3dgs} and emerging 2DGS~\cite{2dgs}.
For clarity, we name 3DGS-based version \emph{\textbf{Trim3DGS}} and 2DGS-based version \emph{\textbf{Trim2DGS}}. For Trim2DGS, we also leverage the depth distortion loss~\cite{barron2022mip, sun2022improved} in 2DGS.

\paragraph{Implementation details.}
We implement TrimGS based on the framework of 3DGS~\cite{3dgs} and extend the CUDA kernel to calculate the contribution of each Gaussian for trimming as discussed in \S~\ref{sec:prune}.
To be compatible with the original 3DGS and 2DGS, we perform a fast 7K-iteration optimization based on their pretrained models.
We conduct trimming every 1,000 iterations and we remove 10\% Gaussian with the lowest contribution each time.
Other training hyperparameters are the same with 3DGS and 2DGS.
For mesh extraction, we utilize the truncated signed distance fusion (TSDF) to fuse rendering median depth maps following 2DGS.
All experiments are conducted on NVIDIA 3090 GPUs.

\begin{figure}[t]
    \centering
    \includegraphics[width=\linewidth]{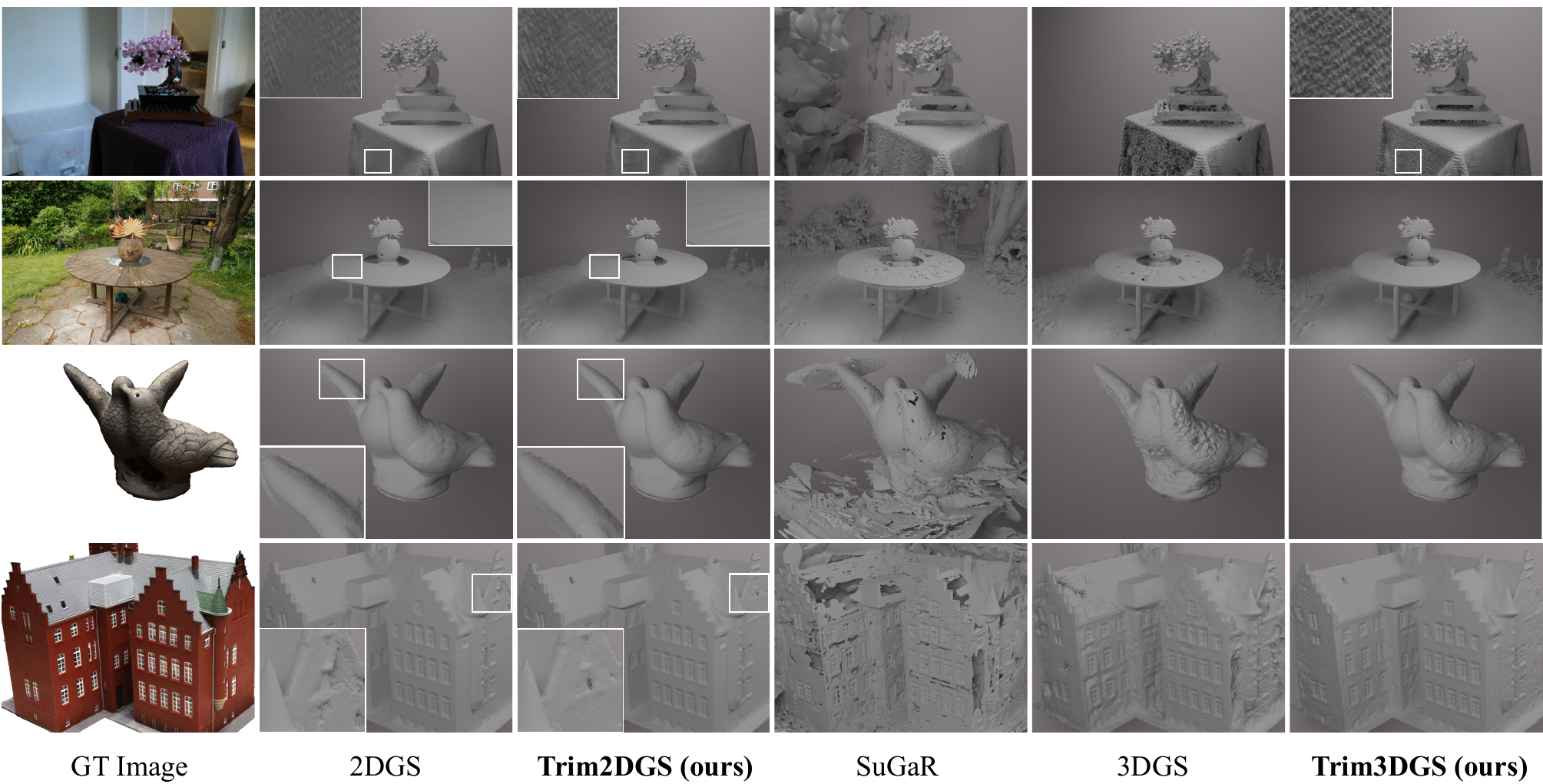}
    \caption{\textbf{Qualitative comparison of meshes on the DTU and MipNeRF360 dataset.} Note SuGaR~\cite{sugar} uses a different mesh extraction strategy from others, so it includes some background. 
    3DGS and SuGaR have \emph{\textbf{messy geometry}} and 2DGS exhibits slight \emph{\textbf{over-smoothness}}.}
    \label{fig:mesh_cmp}
    \vspace{-3pt}
\end{figure}

\setlength\tabcolsep{0.5em}
\renewcommand{\arraystretch}{1.2}
\begin{table*}[ht]
\centering
\caption{\textbf{Quantitative comparison of reconstructed meshes on the DTU Dataset}. }
\resizebox{.98\textwidth}{!}{
\begin{tabular}{@{}llcccccccccccccccclc}
\hline
 \multicolumn{3}{c}{} & 24 & 37 & 40 & 55 & 63 & 65 & 69 & 83 & 97 & 105 & 106 & 110 & 114 & 118 & 122 & & Mean~$\downarrow$  \\ \cline{4-18} \cline{20-20}
\multirow{3}{*}{\rotatebox[origin=c]{90}{implicit}} & NeRF~\cite{mildenhall2021nerf} & & 1.90 & 1.60 & 1.85 & 0.58 & 2.28 & 1.27 & 1.47 & 1.67 & 2.05 & 1.07 & 0.88 & 2.53 & 1.06 & 1.15 & 0.96 & & 1.49 \\
 & VolSDF~\cite{yariv2021volume} & &  1.14 &  1.26 &  0.81 & 0.49 & 1.25 &  \sbest0.70 &  \sbest0.72 &  \tbest 1.29 & \tbest 1.18 &  \sbest 0.70 & \sbest 0.66 & \tbest 1.08 & \tbest 0.42 &  \sbest 0.61 &  0.55 & & 0.86  \\
 & NeuS~\cite{wang2021neus} & & 1.00 & 1.37 & 0.93 &  0.43 & 1.10 &  \best 0.65 &   \best 0.57 &  1.48 &  \best 1.09 &  0.83 &  \best 0.52 &  1.20 & \best 0.35 &  \best 0.49 &  0.54 & &  0.84  \\
 \cline{2-2} \cline{4-18} \cline{20-20}
\multirow{5}{*}{\rotatebox[origin=c]{90}{explicit}} 
&  3DGS~\cite{3dgs} & & 2.14 & 1.53 & 2.08 & 1.68 & 3.49 & 2.21 & 1.43 & 2.07 & 2.22 & 1.75 &  1.79 & 2.55 & 1.53 & 1.52 & 1.50 & & 1.96 \\
 &  SuGaR~\cite{sugar} & & 1.47 & 1.33 & 1.13 & 0.61 & 2.25 & 1.71 & 1.15 & 1.63 & 1.62 & 1.07 & 0.79 & 2.45 & 0.98 & 0.88 & 0.79 & & 1.33 \\
 &GaussianSurfels~\cite{GaussianSurfels}&&0.66&0.93&0.54&\tbest 0.41&\tbest 1.06&1.14&0.85&\tbest 1.29&1.53&0.79&0.82&1.58&0.45&0.66&0.53&& 0.88\\
 & 2DGS~\cite{2dgs} &&  \best 0.48 &  \tbest0.91 &  \sbest 0.39 &  \sbest 0.39 &  \sbest 1.01 &  0.83 &  0.81 &  1.36 &  1.27 &  0.76  & \tbest 0.70 &  1.40 &   \sbest 0.40 &   0.76 &  \tbest 0.52 &&  \tbest 0.80 \\
 & GOF~\cite{gof} & & \sbest 0.50 & \best 0.82 & \best 0.37 & \best 0.37 & 1.12 & \tbest 0.74 & \tbest 0.73 & \best 1.18 & 1.29 & \best 0.68 & 0.77 & \best 0.90 & \tbest 0.42 & 0.66 & \best 0.49 && \sbest 0.74 \\
 & Trim3DGS (ours) & &  \tbest 0.52 & \sbest 0.84 & 0.58 & \tbest 0.41 & 1.07 &  1.02 & 0.82 &\sbest 1.26 & 1.48 & 0.75 & 0.82 & 1.23 & 0.50 &\sbest 0.61 &\tbest 0.52 && 0.83 \\
 & Trim2DGS (ours) & & \best 0.48 &\best 0.82 & \tbest 0.44 &  0.45 & \best 0.95 & 0.75 & 0.74 &\best  1.18 &\sbest 1.13 &\tbest 0.72 & \tbest 0.70 &\sbest 0.99 &\tbest 0.42 &\tbest 0.62 &\sbest 0.50 &&\best 0.72 \\
 \hline
\end{tabular}
}
\label{tab:dtu_mesh}
\vspace{-0.1cm}
\end{table*}

\subsection{Mesh Evaluation}
We first follow the standard evaluation protocol in previous methods~\cite{2dgs} to evaluate the quality of reconstructed meshes on the DTU dataset~\cite{jensen2014large}, as shown in Table~\ref{tab:dtu_mesh}.
Based on the original 3DGS, the proposed Trim3DGS achieves significant performance improvement by reducing Chamfer Distance error by 1.13.
The proposed Trim2DGS also achieves further performance improvement based on the state-of-the-art 2DGS. Figure~\ref{fig:mesh_cmp} shows the qualitative comparison on DTU and MipNeRF-360~\cite{barron2022mip} datasets.

\subsection{Point Evaluation}
\label{sec:point_evaluation}
The TSDF-based mesh extraction adopted in this paper and previous arts can yield smooth meshes.
However, the mesh quality is not the golden standard to evaluate the geometric quality of 3D Gaussian fields for the following two reasons:
(1) The median depth map rendering (Appendix~\ref{sec:appendix_median_depth}) and TSDF generation are quite robust to floaters and noisy Gaussians around the surfaces; 
(2) Due to the limitation of quantization precision in TSDF, geometry details cannot be represented by explicit meshes.

To more properly evaluate the geometric quality of 3D Gaussian fields, we further propose to directly evaluate the quality of raw point clouds (i.e., Gaussian centers) by Chamfer Distance. We adopt a simple downsampling strategy to avoid unfairness caused by different local densities of different Gaussian fields. The details are presented in the Appendix~\ref{sec:appendix_point_eval}.
The results in Table~\ref{tab:dtu_point} show that the proposed method also achieves consistent improvement on the raw point evaluation.
\setlength\tabcolsep{0.5em}
\renewcommand{\arraystretch}{1.2}
\begin{table*}[ht]
\centering
\caption{\textbf{Quantitative comparison of point clouds (i.e., Gaussians' centers) on the DTU dataset}. }
\vspace{-0.2cm}
\resizebox{.98\textwidth}{!}{
\begin{tabular}{@{}lcccccccccccccccclc}
\hline
 \multicolumn{2}{c}{} & 24 & 37 & 40 & 55 & 63 & 65 & 69 & 83 & 97 & 105 & 106 & 110 & 114 & 118 & 122 & & Mean~$\downarrow$  \\ \cline{3-17} \cline{19-19}

3DGS~\cite{3dgs} & & 1.26 & 1.35 & 1.72 & 1.35 & 1.48 & 1.80 & 1.54 & 3.49 & 1.85 & 1.42 &  1.69 & 1.54 & 1.29 & 1.41 & 1.35 & & 1.64   \\
 2DGS~\cite{2dgs} && 0.86 &  1.05 &  0.95 &  0.8 &  1.01 &  1.34 &  1.13 &  3.10 &  1.58 &  1.08  &  1.10 &  1.28 &  0.81 &   0.89 &  0.75 &&  1.18 \\
 \cline{3-17}\cline{19-19}
 Trim3DGS & &  0.97 & 0.94 &  0.97 &  0.73 & 1.21 &  1.54 & 1.27 & 3.46 & 1.72 & 1.30 & 1.14 & 1.40 & 1.10 & 1.14 & 0.97 &&  1.32 \\
 Trim2DGS & & 0.76 & 0.78 &  0.72 &  0.68 & 1.08 &  1.03 & 0.90 &  3.18 & 1.37 & 1.10 & 0.90 & 1.08 & 0.60 & 0.80 & 0.71 && 1.05 \\
 \hline
\end{tabular}
}
\label{tab:dtu_point}
\vspace{-10pt}
\end{table*}

\begin{figure}[t]
    \centering
    \includegraphics[width=0.98\linewidth]{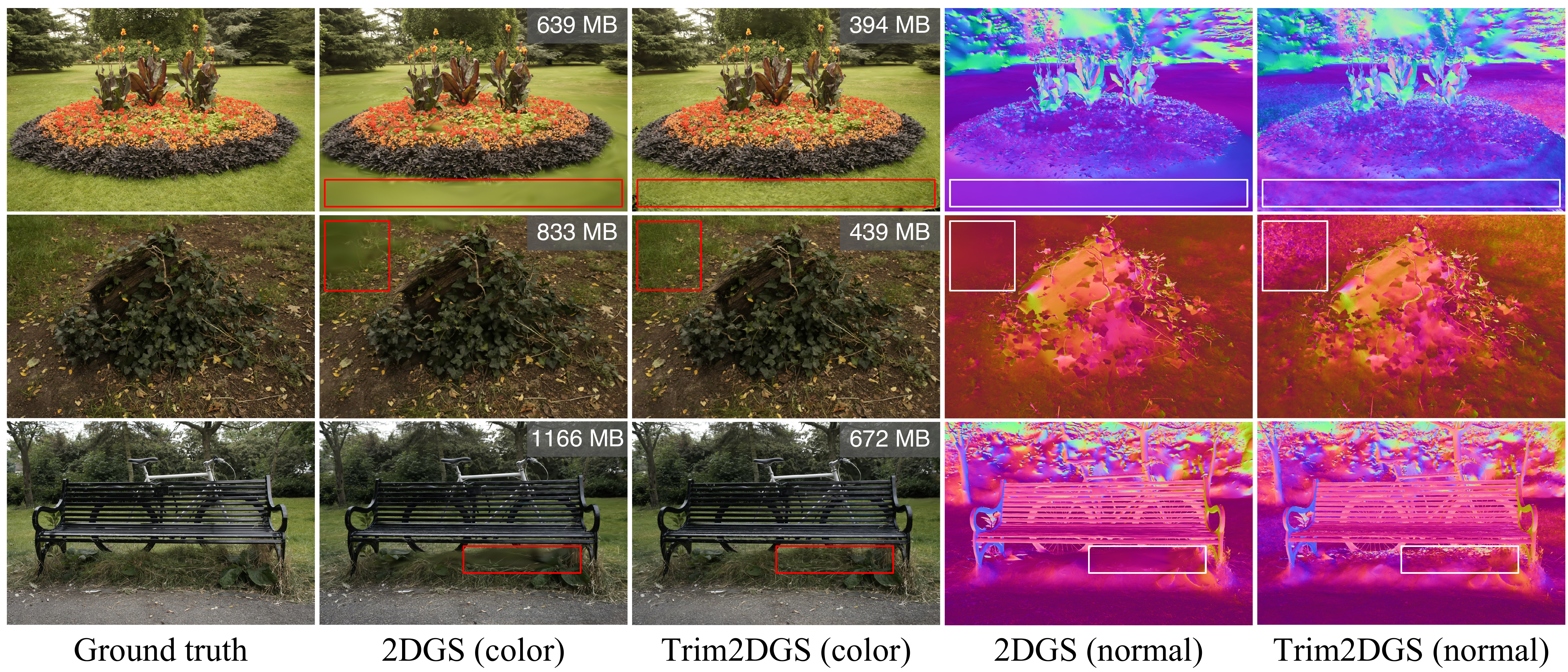}
    \caption{\textbf{Comparison of rendering quality (test-set view) between 2DGS and Trim2DGS in MipNeRF360 dataset.} Our Trim2DGS enhances perceptual quality in high-frequency regions, mitigating the over-smoothness in 2DGS. Notably, Trim2DGS substantially reduces the storage consumption, credited to our proposed contribution-based trimming technique.}
    \label{fig:render_compare}
    \vspace{-3pt}
\end{figure}
\subsection{Rendering Evaluation}
As we discussed in \S~\ref{sec:pre} and \S~\ref{sec:appendix_gradient}, large Gaussians have limited capacity to represent intricate geometry details and high-frequency regions such as \emph{leaves} and \emph{lawn}, which greatly weakens the perceptual quality.
In this sub-section, we demonstrate that our scale control strategy in \S~\ref{sec:split} results in better perceptual rendering quality, as shown in Table~\ref{tab:mean-mipnerf360-scores}.
We emphasize the improvement on the LPIPS metric, which focuses more on the human-vision-sensitive high-frequency and sharp regions as proven in LPIPS~\cite{lpips}. 
In contrast, PSNR is slightly biased towards blurred results due to its least square formulation.

Our visualization results in Figure~\ref{fig:render_compare} also verify this point, where our results show much better rendering quality in high-frequency regions. Since the very tiny geometry details cannot be represented by meshes due to the resolution limitation of TSDF, here we use the normal map to showcase the geometry details. As can be seen, the previous 2DGS tends to \emph{\textbf{over-smooth}} these regions in both RGB images and normals.

We further emphasize that TrimGS does not improve the quality of high-frequency regions by simply splitting more Gaussians, which is shown by the marked storage consumption in Figure~\ref{fig:render_compare}.
This is credited to the proposed contribution-based trimming technique, which accurately identifies and removes the redundant Gaussians.

\begin{table}
\caption{\textbf{Quantitative comparison of rendering quality for MipNeRF360 dataset.} Trim3DGS and Trim2DGS result in better perceptual quality, particularly in terms of the LPIPS metric.
\dag: mean performance of outdoor / indoor scenes. $^*$: our re-implementation.
}
\resizebox{0.98\linewidth}{!}{

\begin{tabular}{l|ccccc|c|cccc|c}
 & bicycle & flowers & garden & stump & treehill & out. mean\dag & room & counter & kitchen & bonsai & in. mean\dag \\
\hline
\emph{\textbf{PSNR}} $\uparrow$ & \\
\hline
3DGS$^*$ & \best 25.18 & \best 21.33 &\best 27.39 & \sbest 26.53 & \best 22.48 & \best 24.58 & \best 31.40 & \best 28.96 & \best 31.22 &\best 32.16 & \best 30.93 \\
SuGaR &  23.12 & 19.25 & 25.43 & 24.69 & 21.33 & 22.76 & 30.12 & 27.57 & 29.48 &  30.59 & 29.44 \\
2DGS$^*$ & 24.75 &\tbest 20.97 &\tbest 26.58 & 26.18 & \tbest 22.31 & \tbest 24.16 & \tbest 30.56 & \tbest 28.01 & \tbest 30.13 & \tbest 31.21 & \tbest 29.98 \\

Trim2DGS & \tbest 24.95 & 20.80 & 26.53 & \tbest 26.28 & 22.01 & 24.11  & 30.29 & 27.91 & 30.03 & 31.12 & 29.84 \\

Trim3DGS & \sbest 25.16 & \sbest 21.29 & \sbest 27.28 & \best 26.59 & \sbest 22.36 & \sbest 24.54 & \sbest 31.05 & \sbest 28.89 & \sbest 31.07 & \sbest 32.03 & \sbest 30.76 \\
\hline
\emph{\textbf{SSIM}} $\uparrow$ & \\
\hline
3DGS$^*$ & \sbest 0.763 & \sbest 0.599 &\best 0.865 & \sbest 0.767 & \best 0.630 &\sbest 0.725&\best 0.919 &\best 0.907 &\sbest 0.923 &\sbest 0.941 & \best 0.923 \\
SuGaR & 0.639 & 0.486 & 0.776 & 0.686 & 0.566 & 0.631 & \tbest 0.910 &\tbest 0.892 & 0.908 & 0.932 &\tbest 0.911 \\
2DGS$^*$ &  0.739 & 0.569 & 0.843 & 0.757 &\tbest 0.621 & 0.706 & 0.905 & 0.890 & 0.914 & 0.929 & 0.910 \\

Trim2DGS & \tbest 0.755 &\tbest 0.580 & \tbest 0.849 & \tbest 0.764 &\sbest 0.622 &\tbest 0.714 &\tbest 0.910 & \sbest 0.896 &\tbest 0.919 &\tbest 0.934 &\sbest 0.915 \\
Trim3DGS & \best 0.767 & \best 0.602 & \sbest 0.864 & \best 0.770 &\best 0.630 & \best 0.727 &\sbest 0.917 &\best 0.907 & \best 0.927 &\best 0.942 &\best 0.923 \\
\hline
\emph{\textbf{LPIPS}} $\downarrow$ & \\
\hline
3DGS$^*$ & \sbest 0.212 &\sbest 0.341 & \best 0.108 &\sbest 0.220 &\tbest 0.330& \sbest 0.242&\tbest 0.220&\sbest 0.201&\sbest 0.128 &\sbest 0.205 &\sbest 0.189 \\
SuGaR & 0.344 & 0.416 & 0.220 & 0.335 & 0.429 & 0.349&0.245 & 0.232 & 0.164 & 0.221 & 0.216 \\
2DGS$^*$ & 0.255 & 0.378 & 0.138 & 0.256 & 0.367 & 0.279& 0.244 & 0.232 & 0.147 & 0.228 & 0.213 \\

Trim2DGS & \tbest 0.216 &\tbest 0.342 &\tbest 0.118 &\tbest 0.229 &\best 0.322 &\tbest 0.246 &\sbest 0.219 &\tbest 0.208 & \tbest 0.133 &\tbest 0.212 &\tbest 0.193 \\
Trim3DGS & \best 0.202 & \best 0.334 & \sbest 0.111 &\best 0.215 &\sbest 0.323 &\best 0.237 &\best 0.218 &\best 0.196 &\best 0.127 &\best 0.201 &\best 0.186 \\
\end{tabular}
}
\label{tab:mean-mipnerf360-scores}
\vspace*{-5pt}
\end{table}
\subsection{Ablation Study}

\paragraph{Effectiveness of major components.}
We first evaluate the effectiveness of major components in TrimGS, including contribution-based trimming, scale control, and normal regularization.
Table~\ref{tab:dtu_roadmap} shows the performance roadmap from the original 3DGS to our full model, leading to the following three findings.
(1) The contribution-based trimming effectively boosts the geometry quality of original 3D Gaussians, especially on point-based evaluation (Table~\ref{tab:dtu_roadmap}). 
(2) Scale control leads to slight improvements. This is because the overall geometric metrics are not sensitive to detail changes and are dominated by smooth surfaces of large areas.
(3) Normal regularization leads to significant improvement for the mesh-based metric but has no help for the point-based metric. This is because normal regularization forces the surfaces to be smoother, therefore having a preference of the mesh quality.
\setlength\tabcolsep{0.5em}
\begin{table*}[th]
\centering
\caption{\textbf{Roadmap from 3DGS to Trim3DGS on the DTU Dataset}.
All experiments are fine-tuned for 7K iterations for fair comparison.}
\resizebox{.98\textwidth}{!}{
\begin{tabular}{@{}llcccccccccccccccclc}
\hline
 \multicolumn{3}{c}{} & 24 & 37 & 40 & 55 & 63 & 65 & 69 & 83 & 97 & 105 & 106 & 110 & 114 & 118 & 122 & & Mean CD~$\downarrow$  \\ 
 \hline
\multirow{4}{*}{\rotatebox[origin=c]{90}{mesh}} &
3DGS finetune 7K & & 1.03 & 0.94 & 1.50 & 0.88 & 2.29 & 1.97 & 1.22 & 1.51 & 1.63 & 1.16 &  1.32 & 1.59 & 1.22 & 0.98 & 0.93 & & 1.35   \\
&+ Trimming & & 1.12 & 0.92 & 1.40 & 0.58 & 2.21 & 1.82 & 1.12 & 1.43 & 1.61 & 1.06 &  1.19 & 1.54 & 0.70 & 0.83 & 0.84 & & 1.22   \\
&+ Scale control & & 1.15 & 0.93 & 1.33 & 0.56 & 2.16 & 1.76 & 1.12 & 1.42 & 1.59 & 1.01 &  1.08 & 1.44 & 0.69 & 0.83 & 0.79 & & 1.19   \\
&+ Normal Reg. & &  0.52 & 0.84 &  0.58 &  0.41 & 1.07 &  1.02 & 0.82 & 1.26 & 1.48 & 0.75 & 0.82 & 1.23 & 0.50 & 0.61 & 0.52 &&  0.83 \\
\hline
\multirow{4}{*}{\rotatebox[origin=c]{90}{point}} &
3DGS finetune 7K & & 1.22 & 1.32 & 1.70 & 1.34 & 1.46 & 1.78 & 1.52 & 3.45 & 1.82 & 1.39 &  1.68 & 1.50 & 1.30 & 1.39 & 1.35 & & 1.61   \\
&+ Trimming && 1.02 & 0.95 & 1.29 & 0.72 & 1.24 & 1.46 & 1.23 & 3.21 & 1.69 & 1.18 & 1.26 & 1.21 & 0.99 & 1.02 & 0.95 && 1.29\\
&+ Scale control & & 1.04 & 0.93 & 1.10 & 0.74 & 1.24 & 1.46 & 1.21 & 3.21 & 1.68 & 1.16 &  1.16 & 1.21 & 1.00 & 1.02 & 0.96 & & 1.27   \\
&+ Normal Reg. & &  0.97 & 0.94 &  0.97 &  0.73 & 1.21 &  1.54 & 1.27 & 3.46 & 1.72 & 1.3 & 1.14 & 1.40 & 1.10 & 1.14 & 0.97 &&  1.32 \\
 \hline
\end{tabular}
}
\label{tab:dtu_roadmap}
\end{table*}

\vspace{-3mm}
\paragraph{Contribution designs.}
Here we adopt some potential alternatives to our default design. (1) The default opacity-based contribution in the original 3DGS. (2) The significance score in LightGaussian~\cite{fan2023light}, based on projection area and Gaussian volume. (3) Unnormalized contribution in Eq.~\eqref{eq:contrib}.
Table~\ref{tab:dtu_contribution} shows that our design outperforms these alternatives, especially on the point-based metric.

Table~\ref{tab:dtu_gamma} shows the effectiveness of different $\gamma$ in Eq.~\eqref{eq:contrib2}. The results demonstrate smaller $\gamma$ has better performance. It reveals that the transmittance term with $(1 - \gamma)$ as exponential is more important than the Gaussian opacity with $\gamma$ as exponential. It again verifies that the proposed contribution is more reasonable than the default opacity-based contribution.
\setlength\tabcolsep{0.5em}
\begin{table*}[th]
\centering
\caption{\textbf{Comparisons with other strategies for contribution evaluation on the DTU dataset.}}
\vspace{-0.2cm}
\resizebox{.98\textwidth}{!}{
\begin{tabular}{@{}llcccccccccccccccclc}
\hline
 \multicolumn{3}{c}{} & 24 & 37 & 40 & 55 & 63 & 65 & 69 & 83 & 97 & 105 & 106 & 110 & 114 & 118 & 122 & & Mean CD $\downarrow$ \\ 
\hline
\multirow{4}{*}{\rotatebox[origin=c]{90}{mesh}} & 
Trim3DGS & &  0.52 & 0.84 &  0.58 &  0.41 & 1.07 &  1.02 & 0.82 & 1.26 & 1.48 & 0.75 & 0.82 & 1.23 & 0.50 & 0.61 & 0.52 &&  0.83 \\
&Opacity-based & & 0.75 & 0.92 & 0.70 & 0.68 & 1.71 & 1.15 & 0.99 & 1.49 & 1.59 & 1.01 & 0.86 &  1.49 & 1.50 & 0.66 & 0.56 & & 1.07   \\
&LightGaussian & & 0.51 & 0.86 & 0.63 & 0.53 & 1.16 & 1.03 & 0.84 & 1.25 & 1.50 & 0.75 & 0.83 &  1.39 & 0.71 & 0.63 & 0.53 & & 0.88   \\
&Unnormalized & & 0.50 & 0.85 & 0.66 & 0.42 & 1.05 & 1.02 & 0.81 & 1.24 & 1.49 & 0.73 & 0.81 &  1.28 & 0.54 & 0.62 & 0.54 & & 0.84   \\
\hline
\multirow{4}{*}{\rotatebox[origin=c]{90}{point}} & Trim3DGS & &  0.97 & 0.94 &  0.97 &  0.73 & 1.21 &  1.54 & 1.27 & 3.46 & 1.72 & 1.3 & 1.14 & 1.40 & 1.10 & 1.14 & 0.97 &&  1.32 \\
&Opacity-based && 1.36 & 1.44 & 1.52 & 1.32 & 2.26 & 2.26 & 1.91 & 4.84 & 2.16 & 2.01 & 1.58 & 1.70 & 2.72 & 1.49 & 1.42 && 2.00\\
&LightGaussian && 1.28 & 1.41 & 1.47 & 1.58 & 1.93 & 2.18 & 1.71 & 3.85 & 1.95 & 1.70 & 1.62 & 2.01 & 2.06 & 1.61 & 1.42 && 1.85\\
&Unnormalized && 1.05 & 1.04 & 1.10 & 0.86 & 1.39 & 1.69 & 1.38 & 3.57 & 1.76 & 1.40 & 1.31 & 1.52 & 1.36 & 1.24 & 1.09 && 1.45\\
 \hline
\end{tabular}
}
\label{tab:dtu_contribution}
\vspace{-10pt}
\end{table*}

\setlength\tabcolsep{0.5em}
\begin{table*}[th]
\centering
\caption{\textbf{Performance of Trim3DGS on the DTU Dataset with different $\gamma$ as discussed in Eq.~\eqref{eq:contrib2}}.}
\vspace{-0.2cm}
\resizebox{.98\textwidth}{!}{
\begin{tabular}{@{}llcccccccccccccccclc}
\hline
 \multicolumn{3}{c}{} & 24 & 37 & 40 & 55 & 63 & 65 & 69 & 83 & 97 & 105 & 106 & 110 & 114 & 118 & 122 & & Mean CD~$\downarrow$  \\
 \hline
\multirow{3}{*}{\rotatebox[origin=c]{90}{mesh}} & $\gamma = 0.25$ & & 0.51 & 0.81 & 0.61 & 0.44 & 1.12 & 1.01 & 0.81 & 1.27 & 1.47 & 0.74 &  0.81 & 1.18 & 0.48 & 0.60 & 0.51 & & 0.82   \\
&$\gamma = 0.50$ & & 0.52 & 0.84 & 0.58 & 0.41 & 1.07 & 1.02 & 0.82 & 1.26 & 1.48 & 0.75 &  0.82 & 1.23 & 0.50 & 0.61 & 0.52 & & 0.83   \\
&$\gamma = 0.75$ & & 0.53 & 0.85 & 0.65 & 0.46 & 1.11 & 1.01 & 0.84 & 1.25 & 1.49 & 0.74 &  0.82 & 1.26 & 0.67 & 0.64 & 0.54 & & 0.86   \\
 \hline
\multirow{3}{*}{\rotatebox[origin=c]{90}{point}} &$\gamma = 0.25$&&0.90 & 0.92 & 1.03 & 0.75 & 1.17 & 1.54 & 1.21 & 3.40 & 1.67 & 1.27 & 1.44 & 1.40 & 1.06 & 1.13 & 0.95& & 1.30\\
& $\gamma = 0.50$ & &0.97 & 0.94 &  0.97 &  0.73 & 1.21 &  1.54 & 1.27 & 3.46 & 1.72 & 1.30 & 1.14 & 1.40 & 1.10 & 1.14 & 0.97 &&  1.32 \\
& $\gamma = 0.75$ & &1.10 & 1.11 & 1.15 & 0.80 & 1.32 & 1.68 & 1.44 & 3.55 & 1.79 & 1.45 & 1.35 & 1.52 & 1.38 & 1.24 & 1.15 && 1.47 \\
\hline
\end{tabular}
}
\label{tab:dtu_gamma}
\vspace{-8pt}
\end{table*}

\paragraph{Scale control.}
\begin{figure}[t!]
    \centering
    \includegraphics[width=\textwidth]{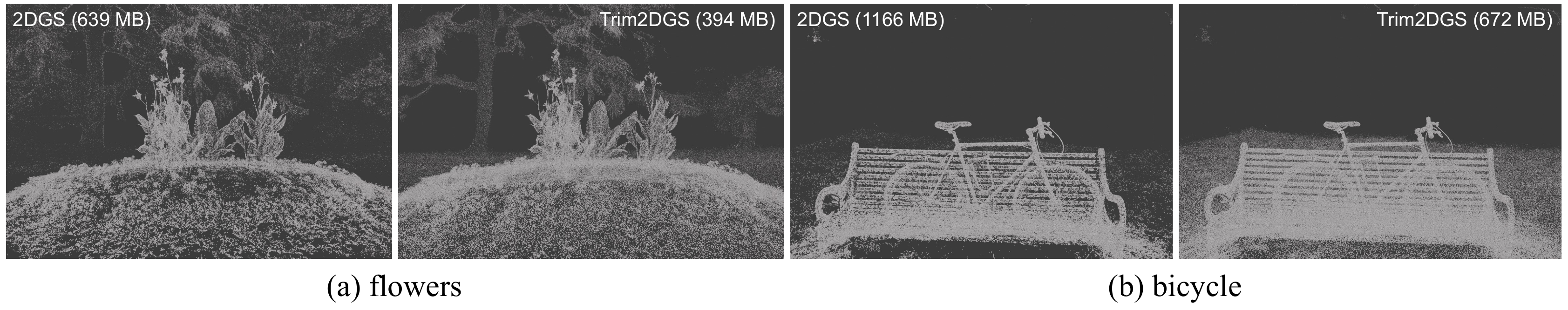}
    \caption{\textbf{Visualization of Gaussian centers between 2DGS and Trim2DGS from \textit{flowers, bicycle} scenes in MipNeRF360 dataset.} Trim2DGS exhibits a more uniform Gaussians while significantly reducing storage consumption. A better video illustration can be found in our project page.}
    \label{fig:mip_pcd}
\end{figure}
As mentioned above, geometric evaluation is not sensitive to detail changes. However, we find perceptual rendering quality is a more sensitive metric.
In Table~\ref{tab:mipnerf_scale_ablation}, Trim2DGS without scale control (\S~\ref{sec:split}) has significant performance drops in terms of the perceptual metric LPIPS.
The qualitative results in Figure~\ref{fig:render_compare} confirm this.
We further visualize Gaussian centers in MipNeRF360 dataset to demonstrate that our Trim2DGS can yield a more uniform distribution of Gaussians while significantly reducing the storage consumption, as can be seen in Figure~\ref{fig:mip_pcd}.
\begin{table}[hbt]
\vspace{-2mm}
\caption{\textbf{Effectiveness of scale control on MipNeRF360.} Scale control improves performance, particularly in terms of the perceptual metric LPIPS, which is instrumental in enhancing the details.}
\resizebox{1.0\linewidth}{!}{
\begin{tabular}{l|ccccc|c|cccc|c}
 & bicycle & flowers & garden & stump & treehill & mean & room & counter & kitchen & bonsai & mean\\
\hline
\emph{\textbf{PSNR}} $\uparrow$ & \\
\hline
2DGS &  24.75 & 20.97 & 26.58 &  26.18 &  22.31 &  24.16 &  30.56 &  28.01 &  30.13 &  31.21 &  29.98 \\
Trim2DGS &  24.95 & 20.80 & 26.53 &  26.28 & 22.01 & 24.11  & 30.29 & 27.91 & 30.03 & 31.12 & 29.84 \\
Trim2DGS (w/o. scale control) & 24.82 & 20.91 & 26.59 & 26.18 & 22.14 & 24.13 & 30.62 & 27.99 & 30.02 & 31.03 & 29.92\\

\hline
\emph{\textbf{SSIM}} $\uparrow$ & \\
\hline

2DGS &  0.739 & 0.569 &  0.843 &  0.757 & 0.621 &  0.706 & 0.905 & 0.890 & 0.914 & 0.929 & 0.910 \\

Trim2DGS &  0.755 & 0.580 &  0.849 &  0.764 & 0.622 & 0.714 & 0.910 & 0.896 & 0.919 & 0.934 & 0.915 \\

Trim2DGS (w/o. scale control) &0.742 &0.572 &0.843&0.757&0.621&0.707&0.906&0.887&0.911&0.927&0.908\\

\hline
\emph{\textbf{LPIPS}} $\downarrow$ & \\
\hline

2DGS & 0.255 & 0.378 & 0.138 & 0.256 & 0.367 & 0.279& 0.244 & 0.232 & 0.147 & 0.228 & 0.213 \\

Trim2DGS &  0.216 & 0.342 & 0.118 & 0.229 & 0.322 & 0.246 & 0.219 & 0.208 & 0.133 & 0.212 & 0.193 \\
Trim2DGS (w/o. scale control) &0.249 &0.376 &0.137&0.253&0.364&0.276&0.246&0.238&0.155&0.249&0.218\\

\hline
\end{tabular}
}
\label{tab:mipnerf_scale_ablation}
\vspace*{-10pt}
\end{table}



\section{Conclusion and Limitations}
\paragraph{Conclusion.}
In this paper, we propose TrimGS to gradually trim 3D Gaussian fields for accurate geometry representation.
It features a new Gaussian contribution evaluation method and a contribution-based trimming strategy.
We also conduct experimental and theoretical analyses and find that large Gaussians are hard to optimize and have limited capacities to represent geometry and appearance details.
So we combine the Trimming strategy with scale control and normal regularization, achieving consistent improvement in geometry reconstruction and perceptual rendering quality.
\vspace{-3mm}
\paragraph{Limitations.}
Although TrimGS emphasizes Gaussian trimming, it still needs geometry regularization terms such as normal consistency.
The geometry regularization inevitably causes a slight drop in rendering quality compared with the original 3DGS, especially for PSNR metric in outdoor scenes as shown in Table~\ref{tab:mean-mipnerf360-scores}.
The rendering quality degradation is also observed in previous 3DGS-based reconstruction methods.
This phenomenon reveals that it remains a challenge to maintain both high rendering and accurate geometry structures.
We will explore this topic in our future work.

{
\small
\bibliographystyle{plainnat}
\bibliography{egbib}
}


\newpage
\appendix

\section{Appendix / supplemental material}

\subsection{Theoretical Analysis for Gradients}\label{sec:appendix_gradient}
\begin{figure}[h]
    \centering
    \includegraphics[width=0.98\linewidth]{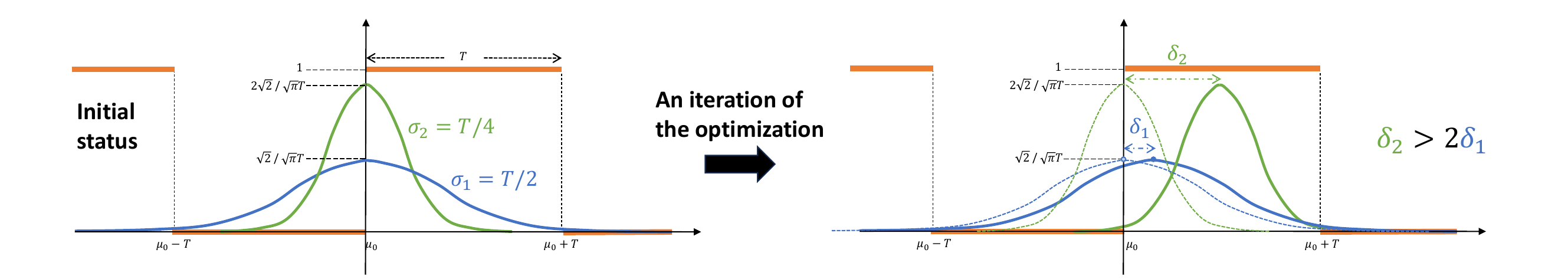}
    \caption{\textbf{A square wave function is locally approximated by Gaussian distributions}. The one with $\sigma=T/4$ gets a larger gradient than the one with $\sigma=T/2$, resulting in a faster move towards the optimal position.}
    \vspace{-3mm}
    \label{fig:gs_square}
\end{figure}

In this section, the gradient dilemma discussed in \S~\ref{sec:prec_grad} is demonstrated in 1D space, where a common square wave function is approximated by 1D Gaussian distributions as shown in Figure~\ref{fig:gs_square}. Let 

\begin{equation}
    f(x)=\left\{
        \begin{aligned}
        &1 \quad x\in[2kT, (2k+1)T) \\
        &0 \qquad \text{otherwise} \\
        \end{aligned}
        \right
        .
\end{equation}

be a square wave function with the period 2T, and $g_{\sigma}(x;\mu)$ be a 1D Gaussian distribution with changeable mean $\mu$ and fixed standard deviation $\sigma$. Considering $L_{\sigma}(\mu) = \int_{-\infty}^{+\infty} |g_{\sigma}(x;\mu) - f(x)| dx$, the $L_1$ loss between $g_{\sigma}$ and $f$, we propose that $g_{\sigma}(x;\mu)$ with relatively smaller $\sigma$ helps with faster convergence when solving the optimal $\mu$ that minimized $L_{\sigma}(\mu)$ by gradient descent. Specifically, with $\sigma_1=T/4$ and $\sigma_2=T/2$, the gradient relationship 
 $|L'_{\sigma_1}(\mu_0)| > 2 |L'_{\sigma_2}(\mu_0)|$ holds when the initial value $\mu_0=2kT$ is set at the start of any pulses of the square wave function.

We prove this by simplifying the expression of the gradient. For convenience, we only consider the finite interval $[\mu_0 - 3\sigma, \mu_0 + 3\sigma]$ due to the $3\sigma$ principle (which is also applied in the 3DGS kernel). Let $I_+$ and $I_-$ denote intervals where $f$ equals 1 and 0 respectively. The gradient can be written as
\begin{equation}
    \begin{aligned}
        L'_{\sigma}(\mu) = \frac{\partial L_{\sigma}(\mu)}{\partial \mu} =& \frac{\partial}{\partial \mu} \int_{\mu_0 - 3\sigma}^{\mu_0 + 3\sigma} |g_{\sigma}(x;\mu) - f(x)| dx \\
        \xlongequal{\text{differentiability of }g}& \int_{\mu_0 - 3\sigma}^{\mu_0 + 3\sigma} \frac{\partial |g_{\sigma}(x;\mu) - f(x)|}{\partial \mu} dx \\
        =& \int_{I_-} \frac{\partial g_{\sigma}(x;\mu)}{\partial \mu} dx - \int_{I_+} \frac{\partial g_{\sigma}(x;\mu)}{\partial \mu} dx .
    \end{aligned}
\end{equation}
Noticing the fact that $\frac{\partial g_{\sigma}(x;\mu)}{\partial \mu} = -\frac{\partial g_{\sigma}(x;\mu)}{\partial x}$ we further conclude that
\begin{equation}\label{eq:newton_leibniz}
    \begin{aligned}
    \frac{\partial L_{\sigma}(\mu)}{\partial \mu} =& -\int_{I_-} \frac{\partial g_{\sigma}(x;\mu)}{\partial x} dx + \int_{I_+} \frac{\partial g_{\sigma}(x;\mu)}{\partial x} dx \\
    \xlongequal{\text{Newton-Leibniz}}& g_{\sigma}(x;\mu) \Big|_{I_+} - g_{\sigma}(x;\mu) \Big|_{I_-} .
    \end{aligned}
\end{equation}

Eq.~\eqref{eq:newton_leibniz} means that the gradient is obtained by adding and subtracting the values of Gaussian distribution at endpoints of intervals. When $\sigma$ is small, these values are more likely to have the same effects on the final gradient (see Eq.~\eqref{eq:same_effects}); when $\sigma$ grows larger, more intervals are incorporated and their effects may be contradictory (Eq.~\eqref{eq:contradictory_effects}).

Specifically, setting $\mu = \mu_0$, when $\sigma = T/4$, we have
\begin{equation}\label{eq:same_effects}
    \begin{aligned}
        &g_{T/4}(x;\mu_0) \Big|_{I_+} - g_{T/4}(x;\mu_0) \Big|_{I_-} \\
        = &(g_{T/4}(\mu_0;\mu_0 + 3T/4) - g_{T/4}(\mu_0;\mu_0)) - (g_{T/4}(\mu_0;\mu_0) - g_{T/4}(\mu_0;\mu_0 - 3T/4)) \\
        = &2(e^{-\frac{9}{2}} -1)/(\sqrt{2\pi}T/4) .
    \end{aligned}
\end{equation}
When $\sigma = T/2$, we have
\begin{equation}\label{eq:contradictory_effects}
    \begin{aligned}
        & g_{T/2}(x;\mu_0) \Big|_{I_+} - g_{T/2}(x;\mu_0) \Big|_{I_-} \\
        = &(g_{T/2}(\mu_0;\mu_0 - T) - g_{T/2}(\mu_0;\mu_0 - 3T/2) + g_{T/2}(\mu_0;\mu_0 + T) - g_{T/2}(\mu_0;\mu_0)) \\
        &- (g_{T/2}(\mu_0;\mu_0) - g_{T/2}(\mu_0;\mu_0 - T) + g_{T/2}(\mu_0;\mu_0 + 3T/2) - g_{T/2}(\mu_0;\mu_0 + T)) \\
        = &2(e^{-2} -1)/(\sqrt{2\pi}T/2) .
    \end{aligned}
\end{equation}
Therefore, $|L'_{T/4}(\mu_0)| > 2 |L'_{T/2}(\mu_0)|$, which indicates that the Gaussian distribution with smaller $\sigma$ gets larger gradient in this situation. 

\section{Implementation}
\subsection{Median Depth Rendering}
\label{sec:appendix_median_depth}
We follow 2DGS to render depth maps.
Specifically, during volume rendering, we hypothesize that the furthest visible Gaussian is closest to the actual surface.
Thus we record the depth of the Gaussian with the largest z-axis value in camera coordinates, whose transmittance $T_i$ exceeds 0.5, as the median depth
\begin{equation}
d = \max_{i\in \{T_i>0.5\}} z_i.
\end{equation}
\subsection{Point Cloud Evaluation}
\label{sec:appendix_point_eval}
As we discussed in \S~\ref{sec:point_evaluation}, we propose an evaluation pattern for the quality of raw point clouds (Gaussian centers) through Chamfer Distance.
Specifically, our evaluation starts with the voxelization of the raw point clouds.
Within each voxel, we retain the point that is closest to the voxel's centroid, discarding the others.
The method ensures a uniform distribution of point clouds, mitigating the bias that may arise from varying local densities across different Gaussian fields.
Following voxelization, we leverage the ground truth point cloud provided by the DTU to compute the Chamfer Distance for the downsampled point cloud, which serves as a metric for evaluating the quality of the Gaussian centers point cloud.


\end{document}